\documentclass[journal,onecolumn,oneside]{IEEEtran}

\ifCLASSOPTIONonecolumn
	\usepackage{setspace} 
\fi

\usepackage[pdftex]{graphicx}
\DeclareGraphicsExtensions{.pdf,.jpeg,.png,.eps}
\usepackage{xcolor}
\usepackage{epstopdf}
\usepackage[caption = false,font = footnotesize]{subfig}
\usepackage[bottom]{footmisc}
\usepackage{cite}
\usepackage{booktabs}
\usepackage{enumitem}% http://ctan.org/pkg/enumitem

\usepackage[cmex10]{amsmath}
\usepackage{amsfonts}
\usepackage{amsthm}
\usepackage{amssymb}
\newtheorem{theorem}{Theorem}
\newtheorem{proposition}[theorem]{Proposition}

\usepackage{setspace}

\usepackage{url}

\newcommand{\vect}[1]{\mathbf{#1}}

\DeclareMathSymbol{\R}{\mathalpha}{AMSb}{"52}

\newcommand{\norm}[2][2]{\left\lVert #2 \right\rVert_{#1}}
\usepackage{bm}
\usepackage{booktabs}
\usepackage{algorithmic}
\usepackage{flushend}
\usepackage{array}

\newcommand{\ra}[1]{\renewcommand{\arraystretch}{#1}}

\newsavebox{\ieeealgbox}
\newenvironment{boxedalgorithmic}
  {\begin{lrbox}{\ieeealgbox}
  	
   \begin{minipage}{\dimexpr\columnwidth-2\fboxsep-2\fboxrule}
   \begin{algorithmic}}
  {\end{algorithmic}
   \end{minipage}
   \end{lrbox}\noindent\fbox{\usebox{\ieeealgbox}}}

\begin{document}

\title{Stochastic Training of Neural Networks via Successive Convex Approximations}

\author{Simone~Scardapane,~\IEEEmembership{Member,~IEEE}
		and Paolo~Di~Lorenzo,~\IEEEmembership{Member,~IEEE}% <-this % stops a space
\thanks{S. Scardapane is with the Department of Information Engineering, Electronics and Telecommunications (DIET), Sapienza University of Rome, Via Eudossiana 18, 00184, Rome. P. Di Lorenzo is with the Department of Engineering, University of Perugia, Via G. Duranti 93, 06125, Perugia, Italy. Email: simone.scardapane@uniroma1.it, paolo.dilorenzo@unipg.it.% <-this % stops a space

Corresponding author: S. Scardapane (email: simone.scardapane@uniroma1.it).}
\thanks{Manuscript received July 11, 2017}}

\markboth{Preprint submitted to IEEE Transactions Neural Networks and Learning Systems}%
{Scardapane \MakeLowercase{\textit{et al.}}: Training Neural Networks with Stochastic SCA}

\maketitle

\begin{abstract}
This paper proposes a new family of algorithms for training neural networks (NNs). These are based on recent developments in the field of non-convex optimization, going under the general name of successive convex approximation (SCA) techniques. The basic idea is to iteratively replace the original (non-convex, highly dimensional) learning problem with a sequence of (strongly convex) approximations, which are both accurate and simple to optimize. Differently from similar ideas (e.g., quasi-Newton algorithms), the approximations can be constructed using only first-order information of the neural network function, in a stochastic fashion, while exploiting the overall structure of the learning problem for a faster convergence. We discuss several use cases, based on different choices for the loss function (e.g., squared loss and cross-entropy loss), and for the regularization of the NN's weights. We experiment on several medium-sized benchmark problems, and on a large-scale dataset involving simulated physical data. The results show how the algorithm outperforms state-of-the-art techniques, providing faster convergence to a better minimum. Additionally, we show how the algorithm can be easily parallelized over multiple computational units without hindering its performance. In particular, each computational unit can optimize a tailored surrogate function defined on a randomly assigned subset of the input variables, whose dimension can be selected depending entirely on the available computational power.
\end{abstract}


\IEEEpeerreviewmaketitle

\section{Introduction}
\label{sec:intro}

\IEEEPARstart{T}{raining} a neural network (NN) involves the minimization of a high-dimensional, non-convex loss function, whose gradients can further exponentially vanish or explode during training \cite{goodfellow2016deep}. Until very recently, this combination appeared too challenging to be confronted directly, and the first wave of deep NNs were trained with expensive, layer-wise, greedy initialization procedures. Today, the general consensus is that poor minima are less of a problem than what previously believed (as opposed to, e.g., pathological saddle points \cite{dauphin2014identifying}). Optimization tools have equivalently shifted to fully supervised routines, employing stochastic first-order algorithms, such as Adam \cite{kingma2014adam}, supplemented by advanced regularization methods, including dropout, batch normalization, and several others. As a result, stochastic gradient descent (SGD) has become a fundamental mainstay in machine learning \cite{hardt2016train}.

To speed up the training process, several researchers have considered the inclusion of curvature information in the optimization step \cite{roux2010fast,martens2010deep,martens2011learning,sohl2014fast}, when working with mini-batches of the full training set. This type of second-order methods (e.g., quasi-Newton) looks for a descent direction by minimizing a quadratic approximation to the cost function \cite{boyd2004convex}. In a stochastic setting, the relative Hessian information is estimated with a noisy version computed from the current batch of data points. However, this quadratic approximation can be unreliable, with the error due to stochasticity compounding and making the overall process hard to implement in a straightforward fashion. As a consequence, these algorithms have failed to gain wide recognition in subsequent years.

This dichotomy between SGD and second-order methods is not limited to the training of NNs, and several methods have been proposed in the optimization literature to overcome it. A very recent proposal in this sense is called the successive convex approximation (SCA) framework \cite{facchinei2015parallel}, which was initially proposed in the context of multi-agent systems \cite{scutari2014decomposition}. Like second-order methods, SCA algorithms work by solving a series of approximations of the original optimization task, called surrogate problems. Differently from Newton techniques, however, these surrogates are not limited to simple quadratic approximations. In fact, any knowledge about the structure of the objective function can be exploited in order to design surrogates which as a result are (in general) both simple to compute and efficient to solve, as long as they maintain the first-order information of the original function (see Section \ref{sec:stochastic_sca_optimization} for more details). Recently, these techniques were also extended to a stochastic setting \cite{mokhtari2017largescale}.

In light of the above, the aim of this paper is to explore stochastic SCA algorithms for training NN models. Specifically, we show that the structure of the optimization problem can be exploited to design particularly efficient surrogate problems, which are obtained by combining a linearization of the original NN model with the (convex) loss function, e.g., the squared loss or the cross-entropy loss. The resulting surrogates can be computed efficiently starting from mini-batches of data, without the need to compute expensive second-order information of the cost function. We focus specifically on the case of the squared loss, giving rise to quadratic optimization problems that can be solved immediately in closed form. As shown in the experimental section, they are able to provide excellent descent directions, surpassing most state-of-the-art stochastic solvers, including Adam \cite{kingma2014adam} and Adagrad \cite{bottou2016optimization}. This is particularly useful in situations where we need fast convergence in a small number of iterations, such as in federated learning environments \cite{brendan2016communication}.
\subsubsection*{Contribution of the paper}
We introduce a general framework for optimizing NN models using stochastic SCA algorithms, and we customize it for different loss functions, such as the squared loss, and different regularizers, including $\ell_2$ and $\ell_1$ norms to promote sparsity. We also consider the case of a non-convex regularization term, and show an immediate extension of the framework to handle it. In addition, we further build on the theory of SCA techniques \cite{facchinei2015parallel} to provide a principled way to parallelize the computation (exploiting, e.g., a multi-core architecture), by defining surrogate problems on subsets of the overall vector of parameters, up to one weight per processor. This is useful for NN models comprising a large set of parameters, where different processors are in charge of optimizing separate portions of the network. 

\subsubsection*{Outline of the paper}
We analyze related literature in Section \ref{sec:related_works}. Next, Section \ref{sec:preliminaries} provides a general stochastic SCA algorithm and its convergence properties. In Section \ref{sec:proposed_optimization_approach}, we show how to customize the algorithm to the problem of NN training, and we explore several specializations employing different loss functions and/or regularizers. In Section \ref{sec:parallelizing_surrogate_optimization} we describe how to parallelize the algorithm. Finally, we provide a comprehensive evaluation in Section \ref{sec:experimental_results}, before concluding in Section \ref{sec:conclusions}.

\subsubsection*{Notation}
We denote vectors using boldface lowercase letters, e.g., $\vect{a}$; matrices are denoted by boldface uppercase letters, e.g., $\vect{A}$. All vectors are assumed to be column vectors. The operator $\norm[p]{\cdot}$ is the standard $\ell_p$ norm on an Euclidean space. For $p=2$, it coincides with the Euclidean norm, while for $p=1$ we obtain the Manhattan (or taxicab) norm defined for a generic vector $\vect{v} \in \R^B$ as $\norm[1]{\vect{v}} = \sum_{k=1}^B |v_k|$. Other notation is introduced along the paper when required.

\section{Related works}
\label{sec:related_works}

The idea of successively replacing a non-convex objective with a series of convex approximations is not novel in the optimization literature, and it appears in a wide range of previous approaches, including convex-concave procedures \cite{yuille2003concave} and proximal minimization algorithms \cite{bolte2014proximal}. However, most previous methods imposed stringent conditions on the approximant, such as it being a global upper bound of the original cost (e.g., the SUM algorithm in \cite{razaviyayn2013unified}). The SCA methods that we consider here originated in the context of multi-agent systems \cite{scutari2014decomposition}, and were later extended to deal with general non-convex optimization problems \cite{scutari2014parallel,facchinei2015parallel,scutari2016parallel}. Under this framework, the (convex) approximation is only required to keep the first-order information of the original (non-convex) cost with respect to the current estimate, thus making its definition highly flexible. Additionally, the optimization problems can be easily decomposed into subproblems (see later in Section \ref{sec:parallelizing_surrogate_optimization}), and convergence to a stationary point can be guaranteed under mild conditions. Several extensions were made to the basic framework, most notably SCA techniques for decentralized environments \cite{di2016next,sun2016distributed}, asynchronous processors \cite{cannelli2016asynchronous}, and stochastic updates \cite{mokhtari2017largescale}.

To the best of our knowledge, the only works that applied SCA techniques for training NNs are \cite{di2016parallel,scardapane2017framework}. However, both are specific to a distributed setting with full batch updates, which is not scalable to the training of NNs with a large-scale dataset or with many parameters. The present paper significantly extends \cite{scardapane2017framework} to the case of stochastic updates computed from mini-batches of the training data. More tangentially related to our paper are the investigations in \cite{scardapane2016distributed} and \cite{manno2016convergent}, which applied SCA techniques for training support vector models, always in a batch setting.

More generally, our definition of the surrogate function (see Section \ref{sec:surrogate_definition}) builds on the fact that the non-convexity is specific to the NN model, while the loss and regularization terms are generally convex with respect to the NN output. Historically, the first work to exploit this idea is \cite{bengio2006convex}. In \cite{bengio2006convex}, it is shown that training a NN with a single-hidden layer, where the number of neurons is also learned, is a convex problem with respect to the weights. Using a sparsity penalty, it is possible to replace the original training problem with a series of convex problems where a single hidden unit is incrementally added. Similar ideas are explored in \cite{JMLR:v18:14-546}. Another paper exploiting a related idea is the `mollifying network' presented in \cite{gulcehre2016mollifying}, where the original problem is successively relaxed by convolving it with a mollifier function. Convexity is guaranteed only in the first iteration of training. None of these papers, however, is connected to the SCA techniques described next.

\section{Preliminaries}
\label{sec:preliminaries}

We begin by stating the NN optimization problem in Section \ref{sec:formulation_of_the_problem}. Then, we describe a generic stochastic SCA algorithm for non-convex optimization problems in Section \ref{sec:stochastic_sca_optimization}. As stated before, in order to be applied SCA techniques require the definition of a proper surrogate cost function. How to exploit the structure of our optimization problem to this end, together with several practical use cases, is the topic of the next section.

\subsection{Formulation of the problem}
\label{sec:formulation_of_the_problem}

We want to train a generic NN $f(\vect{w}; \vect{x})$, which takes as input a real-valued vector $\vect{x} \in \R^d$, and outputs a scalar value. We consider networks with a single output mostly for readability, but everything can be extended seamlessly to the case of multiple outputs. The output of the network depends on a set of $Q$ adaptable parameters (e.g., the weights connecting the layers), that we collect in a single vector $\vect{w} \in \R^Q$ to be optimized depending on some training data.  Additionally, note that we consider the NN as a function of its parameters, in order to make the following notation simpler.

The NN can have any number of layers, nonlinearities, etc., as long as the following assumption is satisfied for any possible input vector.

\vspace{0.5em}
\noindent \textbf{Assumption A [On the NN model]:}
\begin{description}
\item[(A1)] $f \in C^1$, i.e., it is continuously differentiable with respect to $\vect{w}$;\smallskip
\item[(A2)] $f$ has Lipschitz continuous gradient, with respect to $\vect{w}$, for some Lipschitz constant $L > 0$, i.e.:\footnote{Unless specified, all gradients in the manuscript are taken with respect to $\vect{w}$ or its current estimate, which is always clear from the context.}
\begin{equation}
\left\lVert\nabla f(\vect{w}_1; \vect{x}) - \nabla f(\vect{w}_2; \vect{x})\right\rVert_2 \le L \bigl\lVert\vect{w}_1 - \vect{w}_2\bigr\rVert_2 \,,
\label{eq:lip_gradient}
\end{equation}
for any $\vect{w}_1, \vect{w}_2 \in \R^Q$.
\end{description}
These are satisfied by most NN models currently used in the literature, the most notable exception being activation functions with a finite number of non-differentiable points such as ReLu neurons \cite{glorot2011deep}, maxout neurons \cite{goodfellow2013maxout}, piecewise linear adaptable functions \cite{agostinelli2014learning}, and a few others. In fact, the aforementioned cases lead to non-convex, non-differentiable neural networks functions that are not currently handled by the SCA method. An interesting future line of research will be to design optimization methods specifically tailored for such important cases.

We are provided with a training dataset of $N$ input/output pairs $\mathcal{S} = \left\{ \vect{x}_i, y_i \right\}$, and the learning task aims at solving the following regularized optimization problem:
\begin{equation}
\underset{\vect{w}}{\min} \;\; U(\vect{w} ) = \frac{1}{N}\sum_{i=1}^N l\bigl(y_i, f(\vect{w}; \vect{x}_i)\bigr) + \lambda \cdot r(\vect{w}) \,,
\label{eq:NN_opt_problem}
\end{equation}
where $l(\cdot, \cdot)$ is a convex, smooth loss function also satisfying the conditions in Assumption A, while $r(\cdot)$ is a (possibly non-smooth) convex regularization term, and $\lambda > 0$ is a user-defined scalar that weights the two terms. Typical losses used in training NNs are the squared error for regression and the cross-entropy loss for classification, while typical regularizers are $\ell_1/\ell_2$ penalties or a combination of them. All loss functions used in practice are convex with respect to their arguments, with the non-convexity of $f$ making the overall problem non-convex. This point will be essential for the development of the algorithm in the following.

A classical approach to solve \eqref{eq:NN_opt_problem} is to use a stochastic, first-order optimization algorithm, where at the $n$th iteration we sample $L$ indexes in $\left\{1, \ldots, N\right\}$, with $L \ll N$. We denote by $\mathcal{I}_n$ the random variable containing the indexes sampled at time $n$, and by $\mathcal{B}_n$ the corresponding mini-batch of elements extracted from our dataset. We update the current estimate $\vect{w}_n$ following (noisy) gradient information from $U(\vect{w})$ \cite{bottou2016optimization}. In particular, using $\mathcal{B}_n$ to compute a noisy gradient $\widetilde{\nabla} U(\vect{w}_n)$ of \eqref{eq:NN_opt_problem}, the simplest possibility is stochastic gradient descent (SGD):
\begin{equation}
\vect{w}_{n+1} = \vect{w}_n - \alpha_n \widetilde{\nabla} U(\vect{w}_n) \,,
\end{equation} 
where $\alpha_n$ is a (generally decreasing) sequence of step-sizes. This basic strategy can be accelerated in a number of ways, including weight-dependent step-sizes, momentum, gradient averaging, and so on. We refer to \cite{bottou2016optimization} for an up-to-date survey on the topic. See \cite{wilson2017marginal} for a recent critique on some of these accelerated methods for training non-convex models.

\subsection{Stochastic SCA optimization}
\label{sec:stochastic_sca_optimization}

By only exploiting first-order information on the cost function $U(\vect{w})$, SGD and its variants can incur in slow convergence speed and, more in general, they do not leverage efficiently all the information contained in the mini-batch $\mathcal{B}_n$ \cite{martens2010deep}. In contrast, Newton and quasi-Newton methods try to add curvature information to the optimization process, by iteratively minimizing a (noisy) quadratic approximation of the cost function. However, as we discussed in the introduction, these methods require several adjustments to be efficient while training NNs \cite{martens2010deep,ngiam2011optimization}, and they have failed to gain widespread adoption. SCA techniques try to surpass these disadvantages by building approximations of the cost function that aim to preserve as much as possible of its `hidden convexity'. In this section we describe the general stochastic SCA technique introduced in \cite{mokhtari2017largescale}, which is used as the building block for the optimization algorithms of the next section.

Simplifying the notation, denote by $l_i(\vect{w}) = l\bigl(y_i, f(\vect{w}; \vect{x}_i)\bigr)$ the $i$th loss term in \eqref{eq:NN_opt_problem} as a function of the NN parameters. Let us consider a surrogate loss $\widetilde{l}_i(\vect{w}; \vect{w}_n)$ of $l_i$ (with respect to the current weights' estimate), having the following properties:

\begin{table}[t]
\ra{1.3}
\centering
\caption{Pseudocode of the stochastic SCA procedure.}
\begin{boxedalgorithmic}[1]\normalsize
\REQUIRE Training set $\mathcal{S}$, step-size sequence $\left\{\alpha_n\right\}_{n=1}^{\infty}$, averaging sequence $\left\{\rho_n\right\}_{n=1}^{\infty}$, damping factor $\tau$, mini-batch size $L$.
\ENSURE Weight vector $\vect{w}^*$ which is a stationary point of \eqref{eq:NN_opt_problem}.
\STATE Initialize $\vect{d}_0 = \vect{0}$ (or in a data-dependent fashion).
\STATE Initialize weights of the network $\vect{w}_0$.
\FOR{$n = 1,2,\ldots$}
\STATE Draw mini-batch $\mathcal{B}_n$ of size $L$.
\STATE Compute optimum of \eqref{eq:surrogate_problem}.
\STATE Compute $\vect{w}_{n+1}$ according to \eqref{eq:w_update}.
\STATE Compute $\vect{d}_{n+1}$ according to \eqref{eq:d_update}.
\ENDFOR
\RETURN Final estimate $\vect{w}_n$.
\end{boxedalgorithmic}\vspace{0.6em}
\label{alg:sca_optimization}
\end{table}

\vspace{0.5em}
\noindent \textbf{Assumption B [On the surrogate function]:}
\begin{description}
\item[(B1)] $\widetilde{l}_i$ is differentiable and convex with respect to $\vect{w}$ anywhere.
\item[(B2)] $\nabla \widetilde{l}_i(\vect{w}; \vect{w}) = \nabla l_i(\vect{w})$ for any choice of $\vect{w}$.
\item[(B3)] $\widetilde{l}_i$ has Lipschitz continuous gradient for some constant $M$.
\end{description}
\noindent Taken together, these assumptions ensure that the convex surrogate loss $\widetilde{l}_i$ keeps the first-order properties of the original non-convex loss $l_i$, while allowing for a much simpler optimization. These conditions are also relatively general, allowing for a lot of flexibility in the design of the surrogate terms. Given the set of indexes $\mathcal{I}_n$ corresponding to the examples in the current mini-batch $\mathcal{B}_n$, the update at time $n$ is made by solving the following (strongly convex) surrogate optimization problem:
\begin{align}
\widehat{\vect{w}}_{n+1} = & \underset{\vect{w}}{\arg\min} \Biggl\{  \rho_n \cdot \frac{1}{L} \sum_{i \in \mathcal{I}_n} \widetilde{l}_i(\vect{w}; \vect{w}_n) + \lambda r(\vect{w}) \Biggr. \nonumber \\ & \Biggl. + \left(1 - \rho_n\right)\vect{d}_n^T\left(\vect{w} - \vect{w}_n\right) 
  + \tau \norm{\vect{w} - \vect{w}_n}^2 \Biggr\} \,,
\label{eq:surrogate_problem}
\end{align}
where:
\begin{itemize}
\item[(i)] $\vect{d}_n \in \R^Q$ is an auxiliary variable updated as a smoothed average of the gradients considered up to time $n$ (see below for the update equation);
\item[(ii)] $\rho_n$ is a time-dependent scalar weighting the information of the current mini-batch with respect to the historical information kept in $\vect{d}_n$; 
\item[(iii)] the last term, with $\tau > 0$, is a proximal component added to ensure that the optimization problem in \eqref{eq:surrogate_problem} is strongly convex ($\tau$ can be set equal to zero if the surrogate $\widetilde{l}_i$, or the regularizer $r$, are already strongly convex).
\end{itemize}
For the moment we consider optimizing the entire vector $\vect{w}$ simultaneously, which is a special case of the algorithm in \cite{mokhtari2017largescale}; we relax this assumption later on in Section \ref{sec:parallelizing_surrogate_optimization} by allowing for parallel updates of sub-blocks of the vector.
Given $\widehat{\vect{w}}_{n+1}$, we update our current estimate with the following convex combination:
\begin{equation}
\vect{w}_{n+1} = \left(1 - \alpha_n \right)\vect{w}_n + \alpha_n \widehat{\vect{w}}_{n+1} \,,
\label{eq:w_update}
\end{equation}
where $\alpha_n$ is the iteration-dependent step-size. Finally, we update the auxiliary variable $\vect{d}_n$ using a similar step:
\begin{equation}
\vect{d}_{n+1} = \left(1 - \rho_n \right)\vect{d}_n + \rho_n \cdot \biggl( \frac{1}{L} \sum_{i \in \mathcal{I}_n} \nabla l_i(\vect{w}_n) \biggr) \,,
\label{eq:d_update}
\end{equation}
where $\rho_n$ is the same scalar value used in the definition of \eqref{eq:surrogate_problem}. This update ensures that the variable asymptotically converges to the gradient of the loss term in \eqref{eq:NN_opt_problem}. The overall algorithm is summarized in Algorithm \ref{alg:sca_optimization}. Convergence to a stationary point of \eqref{eq:NN_opt_problem} is analyzed in the following proposition.
\begin{proposition}
Given assumptions A-B, assume that the step-size/mixing sequences are chosen such that:
\begin{align*}
\text{(i)} & \lim_{n\rightarrow \infty} \alpha_n = 0, \,\, \sum_{n=1}^\infty \alpha_n = \infty, \,\, \sum_{n=1}^\infty \alpha_n^2 < \infty \,. \\
\text{(ii)} & \lim_{n\rightarrow \infty} \rho_n = 0, \,\, \sum_{n=1}^\infty \rho_n = \infty, \,\, \sum_{n=1}^\infty \rho_n^2 < \infty \,. \\
\text{(iii)} & \lim_{n\rightarrow \infty} \alpha_n / \rho_n = 0 \,.
\end{align*}
Additionally, assume that the sequence $\left\{\vect{w}_n\right\}_{n=1}^\infty$ is bounded.\footnote{Note that this condition is trivially satisfied by imposing a finite (but arbitrarily large) box constraint guaranteeing the boundedness of the sequence.} Then, all the conditions in \cite[Theorem 1]{mokhtari2017largescale} are satisfied, and for every limit point generated by Algorithm \ref{alg:sca_optimization}, there exists a subsequence converging to a stationary point of \eqref{eq:NN_opt_problem} almost surely.
\end{proposition}
Condition (iii) above ensures convergence of the auxiliary variable to the real gradient. Note that this type of almost sure convergence in stochastic non-convex settings is relatively rare in the optimization literature, which is an additional benefit of using SCA techniques \cite{mokhtari2017largescale}. To apply Algorithm \ref{alg:sca_optimization}, we need a principled way to construct $\widetilde{l}_i(\cdot;\cdot)$. This is the topic of the next section.
\section{Proposed optimization approach}
\label{sec:proposed_optimization_approach}
\subsection{Definition of the surrogate function}
\label{sec:surrogate_definition}
An immediate way to satisfy Assumption B is to define $\widetilde{l}_i$ as the first-order linearization of $l_i$:
\begin{equation}
\widetilde{l}_i(\vect{w}; \vect{w}_n) = l_i(\vect{w}_n) + \nabla l_i(\vect{w}_n) \bigl( \vect{w} - \vect{w}_n \bigr) \,.
\label{eq:full_linearization}
\end{equation}
By discarding everything except first-order information, the resulting formulation does not have any definite advantage with respect to a variant of SGD. As a concrete example, consider the case of $\ell_2$ regularization $r(\vect{w}) = \frac{1}{2} \norm{\vect{w}}^2$. Since $r(\cdot)$ is strongly convex, we can set $\tau=0$ in \eqref{eq:surrogate_problem}, and we obtain the closed-form solution:
\begin{equation}
\widehat{\vect{w}}_{n+1} = \frac{1}{\lambda}\biggl( \rho_n \frac{1}{L} \nabla \sum_{i \in \mathcal{I}_n} l_i(\vect{w}_n) + \left(1-\rho_n\right)\vect{d}_n  \biggr) \,.
\end{equation}
where $\lambda$ is the user-defined regularization factor from \eqref{eq:NN_opt_problem}. The resulting update resembles a simplified version of Adam \cite{kingma2014adam} that does not take the second-order moment into account. A similar formulation arises when considering $\ell_1$ regularization instead of $\ell_2$. In this case, the optimum of the surrogate problem can be expressed in closed-form with the use of a soft thresholding operator, e.g., see \cite[Eq. (26)]{scardapane2017framework}.

The surrogate function in \eqref{eq:full_linearization} destroys any information about convexity hidden in the cost function. We can do something smarter by noting that each term $l(y_i, f(\vect{w}; \vect{x}_i))$ in \eqref{eq:NN_opt_problem} is given by the composition of a non-convex function, i.e., $f(\vect{w}; \vect{x}_i)$, the NN model, with a convex loss function, i.e., $l(\cdot, \cdot)$. To preserve the convexity of the latter, we take the first-order-linearization of the NN on a single point as:
\begin{equation}
\widetilde{f}_i(\vect{w}; \vect{w}_n) = f(\vect{w}_n; \vect{x}_i) + \vect{J}_{i,n}^T\left( \vect{w} - \vect{w}_n \right) \,,
\label{eq:f_tilde}
\end{equation}
where $\vect{J}_{i,n} = \nabla f(\vect{w}_n;\vect{x}_i)$ is a $Q$-dimensional vector containing the derivatives of the $i$th NN output with respect to the current estimate of the weights. More in general, it will be a matrix with one column per NN output. We refer to this quantity as the weight Jacobian. It is possible to compute it efficiently for the entire mini-batch via a single back-propagation step whose complexity is linear in the number of parameters, e.g., see \cite[Section 5.3.4]{bishop2006pattern}. Our surrogate loss is then defined by combining $\widetilde{f}_i$ with the loss function, as:
\begin{equation}
\widetilde{l}_i(\vect{w}; \vect{w}_n) = l\bigl(y_i, \widetilde{f}_i(\vect{w}; \vect{w}_n)\bigr) \,.
\label{eq:surrogate_with_partial_linearization}
\end{equation}
It is straightforward to show that \eqref{eq:surrogate_with_partial_linearization} satisfies Assumptions B1-B3, i.e., $\widetilde{l}_i(\vect{w} ;\vect{w}_n)$ is a differentiable, convex function that preserves the first order properties of $l_i$ at point $\vect{w}_n$. In particular, convexity follows from the fact that \eqref{eq:surrogate_with_partial_linearization} is given by the composition of a convex function, i.e., $l$, with an affine mapping, i.e., $\widetilde{f}_i$, see \cite[Section 3.2.2]{boyd2004convex}.

In the remainder of the section, we consider some practical examples resulting from specific choices of $l$ and $r$.

\subsection{Example 1: squared loss with $\ell_2$ regularization}
\label{sec:example_1_ridge_regression}

The first practical implementation we discuss is the use of a squared loss function, coupled with an $\ell_2$ regularization:
\begin{equation}
l(a, b) \triangleq \left( a - b \right)^2 , \;\; r(\vect{w}) \triangleq \frac{1}{2} \norm{\vect{w}}^2 \,.
\label{eq:ridge_regression}
\end{equation}
We call this the ridge regression cost in analogy with the linear case. This is the most common way of training neural networks for regression, e.g., see \cite{flunkert2017deepar} for a very recent example. As before, we can set $\tau=0$ thanks to the presence of a strongly convex regularization term. We define for convenience the `residual' terms $r_{i,n}$ as:
\begin{equation}
r_{i,n} = y_i - f(\vect{w}_n; \vect{x}_i) + \vect{J}_{i,n}^T \vect{w}_n \,.
\label{eq:residual}
\end{equation}
After simple algebra manipulations, we can write the optimization problem in \eqref{eq:surrogate_problem} as a quadratic optimization problem:
\begin{equation}
\widehat{\vect{w}}_{n+1}  =  \underset{\vect{w}}{\arg\min}\;\; \biggl\{ \vect{w}^T \Bigl( \vect{A}_n + \lambda\vect{I} \Bigr)\vect{w}_i - 2 \vect{b}_n^T\vect{w} \biggr\} \,,
\label{eq:surrogate_ridge_pl}
\end{equation}
with $\vect{I}$ being the identity matrix of appropriate size, and we defined:
\begin{align}
\vect{A}_n & = \frac{\rho_n}{L}\sum_{i \in \mathcal{I}_n} \vect{J}_{i,n}\vect{J}_{i,n}^T \,, \label{eq:Ai}\\
\vect{b}_n & = \frac{\rho_n}{L}\sum_{i \in \mathcal{I}_n} \vect{J}_{i,n}r_{i,n} - \frac{\left(1-\rho_n\right)}{2} \vect{d}_n \label{eq:bi}\,.
\end{align}
Thus, the solution of the quadratic problem \eqref{eq:surrogate_ridge_pl} is given by:
\begin{equation}
\widehat{\vect{w}}_{n+1} = \Bigl( \vect{A}_n + \lambda\vect{I} \Bigr)^{-1}\vect{b}_n \,.
\label{eq:ridge_PL}
\end{equation}
This requires the inversion of a $Q \times Q$ matrix, which can become impractical for large $Q$. In Section \ref{sec:parallelizing_surrogate_optimization} we show a principled way to decompose this problem and obtain a significant speedup when using multiple processors. Alternatively, one can solve the original problem in \eqref{eq:surrogate_ridge_pl} using, e.g., highly customized conjugate gradient optimization procedures \cite{moller1993scaled}. If we set $\rho_n=1$ and $\mathcal{B}_n = \mathcal{S}$ for any $n$ (i.e., we work in a batch fashion), we recover the PL-SCA algorithm proposed in \cite{scardapane2017framework}, where PL stands for `partial linearization'. For a very small $\lambda$, solving \eqref{eq:ridge_PL} can give rise to numerical problems, because $\vect{A}_n$ has at most rank $L$, being a sum of $L$ rank-1 matrices. In this case, one can set $\tau > 0$, obtaining a slightly modified solution:
\begin{equation}
\widehat{\vect{w}}_{n+1} = \Bigl( \vect{A}_n + \left(\lambda+\tau\right)\vect{I} \Bigr)^{-1}\left(\vect{b}_n + \tau\vect{w}_n\right)\,,
\label{eq:ridge_PL_with_tau}
\end{equation}
\noindent which still guarantees convergence of the stochastic SCA procedure to a local solution of \eqref{eq:NN_opt_problem}. 

\subsection{Case 2: sparsity-inducing penalties}

As a second use case, we consider again the use of a squared loss function, this time combined with an $\ell_1$ regularization to achieve sparsity on the weights:
\begin{equation}
r(\vect{w}) \triangleq \norm[1]{\vect{w}} = \sum_{i=1}^Q \lvert w_i \rvert \,.
\end{equation}
Proceeding as in \eqref{eq:surrogate_ridge_pl}, we can immediately formulate the surrogate problem as an $\ell_1$-regularized quadratic problem:
\begin{equation}
\widehat{\vect{w}}_{n+1}  =  \underset{\vect{w}}{\arg\min}\;\; \biggl\{ \vect{w}^T \vect{A}_n \vect{w}_i - 2 \vect{b}_n^T\vect{w} + \lambda\norm[1]{\vect{w}} \biggr\} \,.
\label{eq:surrogate_lasso_pl}
\end{equation}
There is a wide range of fast solvers for \eqref{eq:surrogate_lasso_pl} \cite{bach2012optimization}, most notably the fast iterative shrinkage and thresholding algorithm (FISTA) \cite{beck2009fast}, which achieves a $\mathcal{O}(1/n^2)$ rate of convergence. Note that this approach will yield exactly sparse solutions at every iteration. On the contrary, it is customary in the NN literature to solve $\ell_1$ regularized problems with SGD algorithms, in which case an exactly sparse solution can never be reached \cite{bengio2012practical}, and a further thresholding step is needed.

This formulation can be extended immediately to similar forms of $\ell_1$ regularization, such as elastic net penalties (given by a weighted sum of $\ell_2$ and $\ell_1$ normalization), and recent group sparse formulations such  as \cite{wen2016learning} and \cite{scardapane2017group}. Group sparse penalties can be used to favor structured forms of sparsity where entire neurons are removed in the optimization process. Suppose the neurons are indexed as $1, \ldots, P$, and denote by $\vect{w}_p \subset \vect{w}$ the set of weights outgoing from the $p$th neuron, such that:
\begin{equation}
\vect{w} = \bigcup_{p=1}^P \vect{w}_p \,.
\end{equation}
Group sparse regularization is achieved as:
\begin{equation}
r(\vect{w}) \triangleq \sum_{p=1}^P a_p \norm{\vect{w}_p} \,,
\label{eq:group_sparse_regularizer}
\end{equation}
where $a_p$ are scalar coefficients defined as the square root of the dimensionality of the corresponding groups. Most optimization algorithms designed to optimize \eqref{eq:surrogate_lasso_pl} can be applied equivalently even if we interchange the $\ell_1$ term with the group sparse term in \eqref{eq:group_sparse_regularizer}. Again, this allows us to obtain exactly sparse solutions.

\subsection{Case 3: cross-entropy loss}

As a third use case, we consider a binary classification problem with $y_i = \left\{0,1\right\}$, for which it is common to optimize the cross-entropy loss defined as:
\begin{equation}
l(a,b) \triangleq a\log(b) + (1-a)\log(1-b) \,.
\label{eq:cross_entropy_loss}
\end{equation}
The previous loss can be combined with either $\ell_2$ or $\ell_1$ regularization, depending on the learning task. Some care must be taken here because, even when the original NN model $f(\cdot)$ is always bounded, the same is not true for its linearization in \eqref{eq:f_tilde}. Substituting \eqref{eq:f_tilde} in \eqref{eq:cross_entropy_loss} is then undefined whenever $\widetilde{f}$ is non-positive or larger than $1$. To solve this issue, note that for binary classification problems with the cross-entropy loss the NN can always be written as:
\begin{equation}
f(\vect{w}) = \sigma\bigl(f^L(\vect{w})\bigr) \,,
\label{eq:f_linear}
\end{equation}
where $\sigma(\cdot)$ is a squashing function (e.g., sigmoid) ensuring that the output is properly defined as a probability, and $f^L$ denotes the output of the NN up to the last nonlinearity. We define a linearization on $f^L$ similar to \eqref{eq:f_tilde}:
\begin{equation}
\widetilde{f}^L_i(\vect{w}; \vect{w}_n) = f^L(\vect{w}_n; \vect{x}_i) + \bigl(\vect{J}^L_{i,n}\bigr)^T\left( \vect{w} - \vect{w}_n \right) \,,
\label{eq:fL_tilde}
\end{equation}
where $\vect{J}^L_{i,n} = \nabla f^L(\vect{w}_n;\vect{x}_i)$. The sigmoid function is neither convex nor concave, but the combination of a sigmoid with \eqref{eq:cross_entropy_loss} is convex. Thus, proceeding as in \eqref{eq:surrogate_with_partial_linearization}, a proper surrogate to use in \eqref{eq:surrogate_problem} is obtained as:
\begin{equation}
\widetilde{l}_i(\vect{w}; \vect{w}_n) = l_i\bigl(y_i, \sigma(\widetilde{f}^L(\vect{w}; \vect{w}_n))\bigr) \,.
\end{equation}
The final optimization problem in \eqref{eq:surrogate_problem} is similar to a standard logistic regression, with the straightforward inclusion of a linear term and a proximal norm on $\vect{w}$. Also here we can exploit highly customized solvers for optimizing the resulting strongly convex problem, e.g., see \cite{yuan2012recent} for a recent survey up to $2012$.

\subsection{Case 4: non-convex regularizers}

The previous sections described use cases that are relatively common in the NN literature. One interesting extension concerns the use of \textit{non-convex} regularization terms. These terms generally arise because of the need to regularize (in some meaningful way) the output of the NN, or the activations of specific neurons. By making both the loss term and the regularization term highly non-convex, they are generally harder to optimize and less common in the literature. However, they fit naturally in the SCA framework, because we can apply the same ideas described in Section \ref{sec:proposed_optimization_approach} to also convexify the regularization term.

As a specific case, consider the use of manifold regularization in deep networks \cite{tomar2014manifold}. The basic idea is to force the NN to provide similar outputs whenever two inputs are `close' according to some distance measure. To this end, suppose that $q_{ij}$ is a non-negative value measuring the distance between the inputs $\vect{x}_i$ and $\vect{x}_j$. Typically, this is defined as some measure of the Euclidean distance for the $k$-nearest neighbors of $\vect{x}_i$, and $0$ otherwise \cite{tomar2014manifold}. A manifold regularization term can then be written as:
\begin{equation}
r(\vect{w}) = \sum_{i=1}^N \frac{1}{k} \sum_{j \in \mathcal{N}_i} q_{ij} \norm{f(\vect{w}; \vect{x}_i) - f(\vect{w}; \vect{x}_j)}^2 \,.
\label{eq:manifold_regularizer}
\end{equation}
Note that, for a mini-batch of elements, computation of \eqref{eq:manifold_regularizer} requires $Lk$ additional forward/backward computations in general \cite{tomar2014manifold}, one for each neighbor of the elements in the mini-batch. We can handle this sort of regularization by replacing $r(\cdot)$ in \eqref{eq:surrogate_problem} with a strongly convex approximation, following the same methodology as in Section \ref{sec:surrogate_definition}. In particular, substituting $f(\cdot)$ with its first-order linearization in \eqref{eq:manifold_regularizer} we can rewrite it as:
\begin{equation}
\widetilde{r}(\vect{w}; \vect{w}_n) = \sum_{i=1}^{N} \frac{1}{k} \sum_{j \in \mathcal{N}_i} q_{ij} \norm{\Delta_{ij} - \vect{J}_{i,j,n}^T\vect{w}}^2 \,,
\end{equation}
where we defined:
\begin{eqnarray}
\Delta_{ij} & = & \bigl(f(\vect{x}_i; \vect{w}_n) - f(\vect{x}_j; \vect{w}_n)\bigr) - \vect{J}_{i,j,n}^T\vect{w}_n \,, \\
\vect{J}_{i,j,n} & = & \left(\vect{J}_{i,n} - \vect{J}_{j,n}\right) \,.
\end{eqnarray}
Thus, manifold regularization provides just a linear and a quadratic term in $\vect{w}$, and can easily be plugged-in with any choice of loss function described before. Particularly, a closed-form solution is preserved when employing the squared loss, whereas a simple addition of a quadratic term is achieved in the case of the cross-entropy loss function.

\section{Parallelizing the surrogate optimization}
\label{sec:parallelizing_surrogate_optimization}
In general, the algorithms described in the previous section are more expensive in computational terms with respect to accelerated gradient methods. This is both their advantage (resulting in faster convergence speed), and their drawback when moving to a large-scale regime. In this section, we show a simple way to parallelize their computation, whenever we have access to a multi-core (or multi-machine) environment.

Roughly, NN training algorithms can be parallelized by partitioning the data (i.e., providing a different mini-batch to each processing unit), or by partitioning the optimization variable $\vect{w}$. For the moment, we focus on the latter strategy. To this end, suppose that $\vect{w}$ is partitioned in $C$ non-overlapping blocks $\vect{w}_{1}, \ldots, \vect{w}_{C}$, so that $\vect{w} = \bigcup_{c=1}^C \vect{w}_{c}$. $\vect{w}_{-c} \triangleq (\vect{w}_{p})_{1=p\neq c}^C$ will denote the tuple of all blocks excepts the $c$-th one, and similarly for all other variables. Additionally, we assume that the regularization term $r$ is block separable, i.e., $r(\vect{w})=\sum_{c=1}^C r_{c}(\vect{w}_{c})$ for some $r_{c}$. This is true for the $\ell_2$ and $\ell_1$ norms, and it holds true also for the group sparse norm in (\ref{eq:group_sparse_regularizer}) if we choose the groups in a consistent way.

At the $n$th iteration each computing unit gets assigned the $c$th block $\vect{w}_{c,n}$ of $\vect{w}_n$. Then, each core solves a smaller surrogate problem defined as:
\begin{align}
\widehat{\vect{w}}_{c,n+1} = & \underset{\vect{w}_c}{\arg\min} \Biggl\{  \rho_n \cdot \frac{1}{L} \sum_{i \in \mathcal{I}_n} \widetilde{l}_{i,c}(\vect{w}_c; \vect{w}_{-c}, \vect{w}_n) + \lambda r_c(\vect{w}_c) \Biggr. \nonumber \\ & \Biggl. + \left(1 - \rho_n\right)\vect{d}_{c,n}^T\left(\vect{w}_c - \vect{w}_{c,n}\right) 
  + \tau \norm{\vect{w}_c - \vect{w}_{c,n}}^2 \Biggr\} \,,
\label{eq:surrogate_problem_parallel}
\end{align}
where $\widetilde{l}_{i,c}(\cdot)$ is a surrogate term respecting assumption B on the block $\vect{w}_c$ only. Each core $c$ can then minimize its corresponding term independently of the others, and their solutions can be aggregated to form the final solution vector. To obtain the surrogate function associated to each core $c$, we simply compute a full surrogate as in the previous section, then fix all the variables $\vect{w}_{-c, n}$ to their current value, such that the resulting function depends only on $\vect{w}_{c}$.

As an example of this idea, consider the ridge surrogate defined in Section \ref{sec:example_1_ridge_regression}. Simple algebra shows that the solution of the local surrogates in a parallel environments are given by:
\begin{equation}
\widetilde{\vect{w}}_{c,n} = \left( \vect{A}_{c,c,n} + \lambda\vect{I} \right)^{-1}\left(\vect{b}_{c,n} - \vect{A}_{c,-c,n}\vect{w}_{-c,n} \right) \,,
\label{eq:rige_parallel_solution}
\end{equation}
where $\vect{A}_{c,c,n}$ is the block (rows and columns) of the matrix $\vect{A}_{n}$ in (\ref{eq:Ai}) corresponding to the $c$-th partition, whereas $\vect{A}_{c,-c,n}$ takes the rows corresponding to the $c$-th partition and all the columns not associated to $c$. Each core has now to invert a matrix having (approximately) size $\frac{1}{C}$ of the original one, thus remarkably reducing the overall computational burden. Similar arguments can be used also to parallelize all the other formulations described in the previous section.

The theorems in \cite{mokhtari2017largescale} ensure that convergence is guaranteed even in the parallel case. Interestingly, convergence is also guaranteed in the more general case where the number of blocks is larger than the number of computational units, and at each iteration every processor is randomly assigned a block \cite{mokhtari2017largescale}. Thus, the size of the blocks (and, consequently, of the surrogate problems to be solved) can be freely chosen based on the available computational requirements. Even more generally, one can consider randomly assigning \textit{both} a block variable and a separate mini-batch. Note, however, that in general every core will have to compute the entire Jacobian matrix in \eqref{eq:f_tilde} due to the back-propagation step. As a matter of fact, this is a general limitation of any parallel gradient-based approach to NN training \cite{seide2014parallelizability}, which is generating new fields of research in term of signal back-propagation, e.g., see \cite{lillicrap2016random}.

\section{Experimental results}
\label{sec:experimental_results}

\begin{table*}[t]
\caption{Schematic description of the mid-sized datasets. For the NN topology, $x/y$ denotes a NN with two layers of dimensions $x$ and $y$ respectively.}
\setlength{\tabcolsep}{4pt}
\renewcommand{\arraystretch}{1.3}
\centering
\begin{small}
\begin{tabular}{@{}p{0.1\columnwidth}p{0.1\columnwidth}p{0.1\columnwidth}p{0.15\columnwidth}p{0.4\columnwidth}p{0.05\columnwidth}@{}}   %@{}p{0.3\columnwidth}p{0.2\columnwidth}p{0.2\columnwidth}@{}
\toprule
\textbf{Dataset} & \textbf{Samples} & \textbf{Features} & \textbf{NN Topology} & \textbf{Prediction task} & \textbf{Source} \\
\midrule
CASP & $45730$ & $9$ & $9/10/6/1$ & Protein's structure prediction from physicochemical properties & UCI \\
Parkinsons & $5875$ & $19$ & $19/15/5/1$ & UPDRS score from measurements of a remote telemonitoring device & UCI \\
SkillCraft1 & $3395$ & $20$ & $18/15/10/1$ & Predicted level from in-game statistics & UCI \\
Wine & $4898$ & $12$ & $11/10/4/1$ & Wine quality from chemical measurements & UCI \\
\bottomrule
\end{tabular}
\end{small}
\label{tab:medium_sized_datasets}
\vspace{0.5em}
\end{table*}

In this section we evaluate the convergence behavior of the proposed technique. We focus on the ridge case considered in Section \ref{sec:example_1_ridge_regression}. We consider several medium-sized datasets in Section \ref{sec:results_medium_sized_datasets}, and a large-scale problem in Section \ref{sec:results_large_scale_dataset}. Python code to repeat the experiments is available under open-source license on the web.\footnote{\url{https://bitbucket.org/ispamm/sca-optimization-for-neural-networks}} Back-propagation and part of the comparing algorithms are based on the AutoGrad library \cite{maclaurin2015gradient}.

For all experiments, the original dataset is normalized so that inputs lie in the $\left[-0.5, 0.5\right]$ range, and outputs lie in the $\left[-0.9, 0.9\right]$ range. For each run, a random $25\%$ of the dataset is used for testing, and the rest for training the network. All experiments are repeated $100$ times by varying the data partitioning and the NN initialization. Missing data is replaced with the median value for the corresponding feature in the entire dataset.

Regarding the NN structure, we use hyperbolic tangent nonlinearities in all neurons, and weights are initialized using the normalized strategy described by \cite{glorot2010understanding}.

\subsection{Experiments on mid-sized datasets}
\label{sec:results_medium_sized_datasets}

We start by consider four mid-sized regression datasets, whose characteristics are briefly summarized in Table \ref{tab:medium_sized_datasets}. All of them were downloaded from the UCI repository.\footnote{\url{https://archive.ics.uci.edu/ml/}} The fourth column in Table \ref{tab:medium_sized_datasets} describe the topology of the NN we have chosen. These parameters are chosen based on an analysis of previous literature in order to obtain state-of-the-art results. However, we underline that our aim is to compare different solvers for the same NN optimization problem, and for this reason only relative differences in accuracy are of concern. For all datasets, we select a small regularization coefficient $\lambda=10^{-3}$, which is found to provide good results.

We compare the results of the algorithm in Section \ref{sec:example_1_ridge_regression} with respect to four state-of-the-art solvers, in terms of mean-squared error (MSE) over the test data, when solving the global optimization problem with the ridge regression cost in \eqref{eq:ridge_regression}. Specifically, we consider the following algorithms:

\begin{description}
	\item[\textbf{Stochastic gradient descent} (GD)]: this is a simple first-order steepest descent procedure with diminishing step-size (see below).
	\item[\textbf{Adagrad}]: differently from SGD, we use different step-sizes per weight, which evolve according to the relative values of the gradients' updates \cite{duchi2011adaptive}.
	\item[\textbf{RMSProp}]: it also considers adaptive independent step-sizes; however they are adapted based on an exponentially-weighted moving average \cite{zeiler2012adadelta}.
	\item[\textbf{Adam}]: Adam combines a momentum strategy with adaptive step-sizes, where both first- and second-order moments are computed in an streaming fashion \cite{kingma2014adam}.
\end{description}

For the proposed algorithm and SGD, we use the following quadratically decreasing rule for selecting the step-sizes:
\begin{equation}
\alpha_n = \alpha_{n-1}\left( 1 - \varepsilon\alpha_{n-1} \right) \,,
\label{eq:step_size_sequence}
\end{equation}
where $\alpha_0$ and $\varepsilon$ are selected by the user. The same sequence is also used for selecting $\rho_n$ in the proposed algorithm. For fairness of comparison, we selected default values for all algorithms leading to (in average) their best convergence behavior. In particular, we set $\alpha_0 = 0.5$, $\varepsilon = 0.01$, $\tau=0$, and $\rho_0 = 0.9$ for SCA, $\alpha_0 = 0.1$, and $\varepsilon = 0.01$ for SGD. For AdaGrad and RMSProp, we set the initial learning rate to $0.01$. For the latter, we set a decaying value of $\gamma=0.9$. For Adam, we use the default values as in \cite{kingma2014adam}. For all algorithms, we consider randomly extracted mini-batches of size $L=20$.

\begin{figure*}[t]
	\centering
	\subfloat[CASP]{\includegraphics[width=0.45\columnwidth]{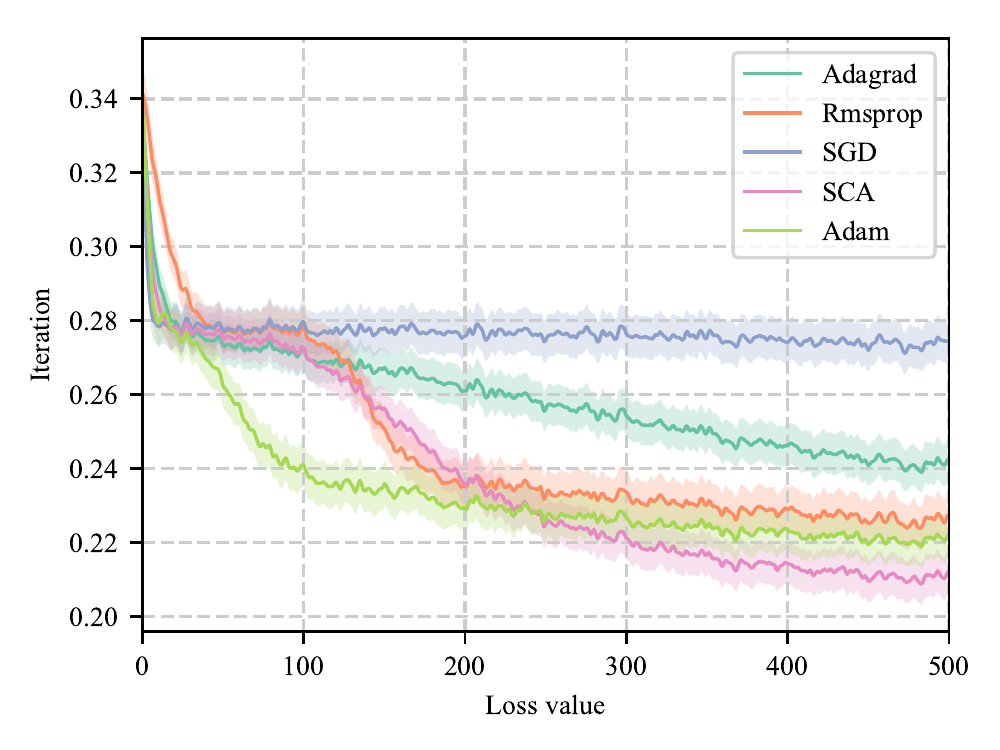}%
		\label{fig:conv_casp}}
	\hfil
	\subfloat[Parkinson]{\includegraphics[width=0.45\columnwidth]{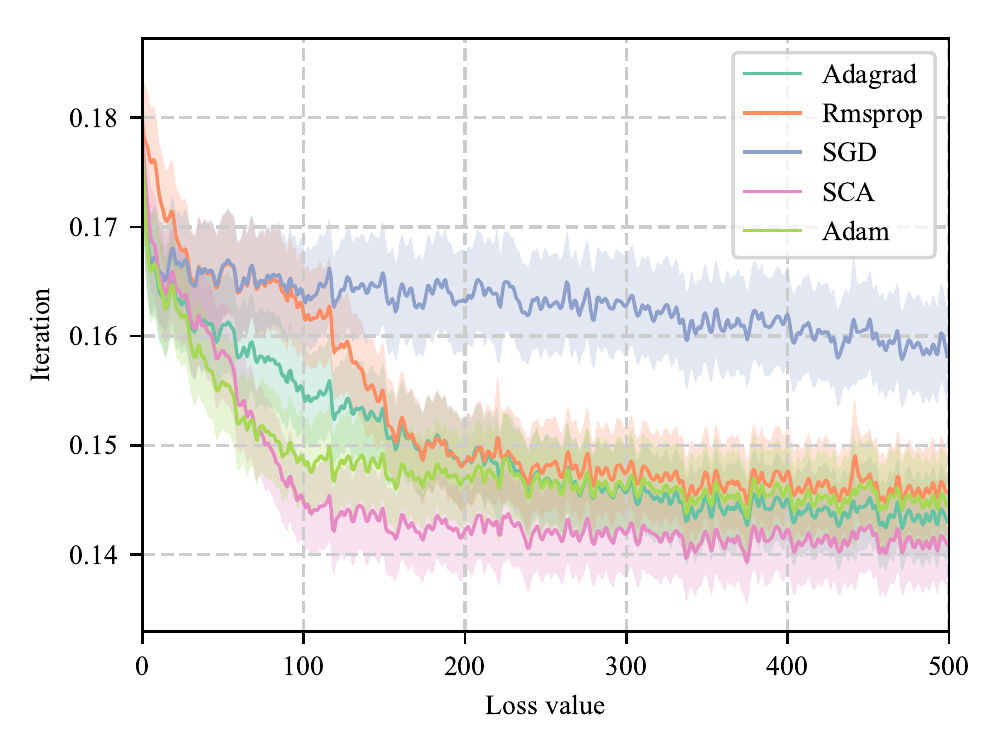}%
		\label{fig:conv_parkinson}}
	\vfill
	\subfloat[Skills]{\includegraphics[width=0.45\columnwidth]{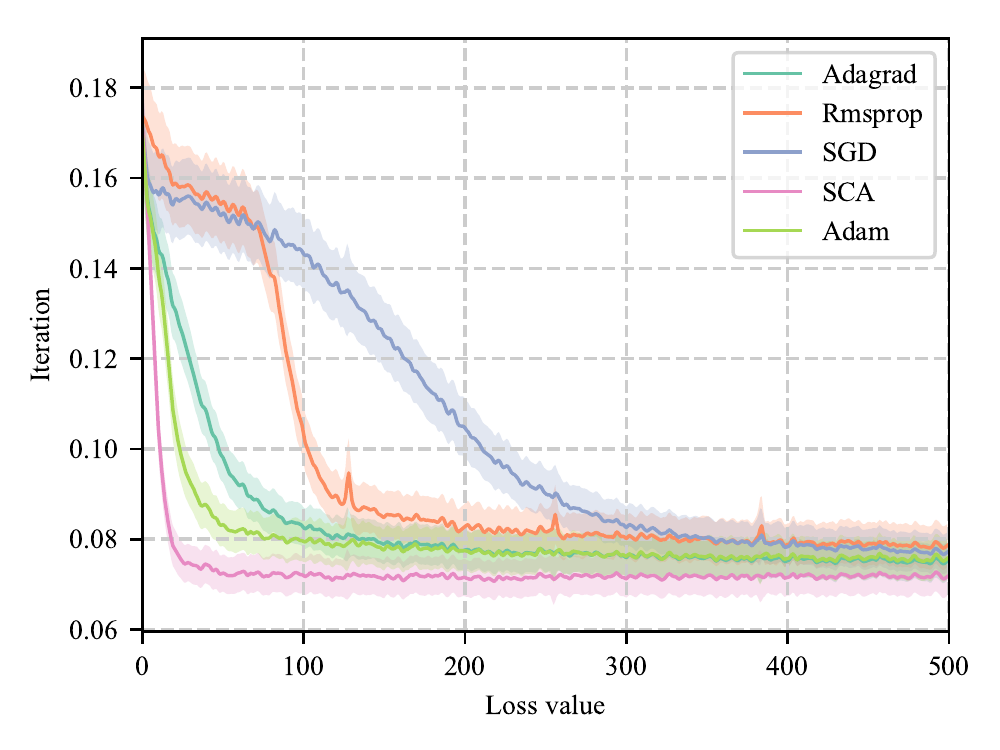}%
			\label{fig:conv_skills}}
	\hfil
	\subfloat[Wine]{\includegraphics[width=0.45\columnwidth]{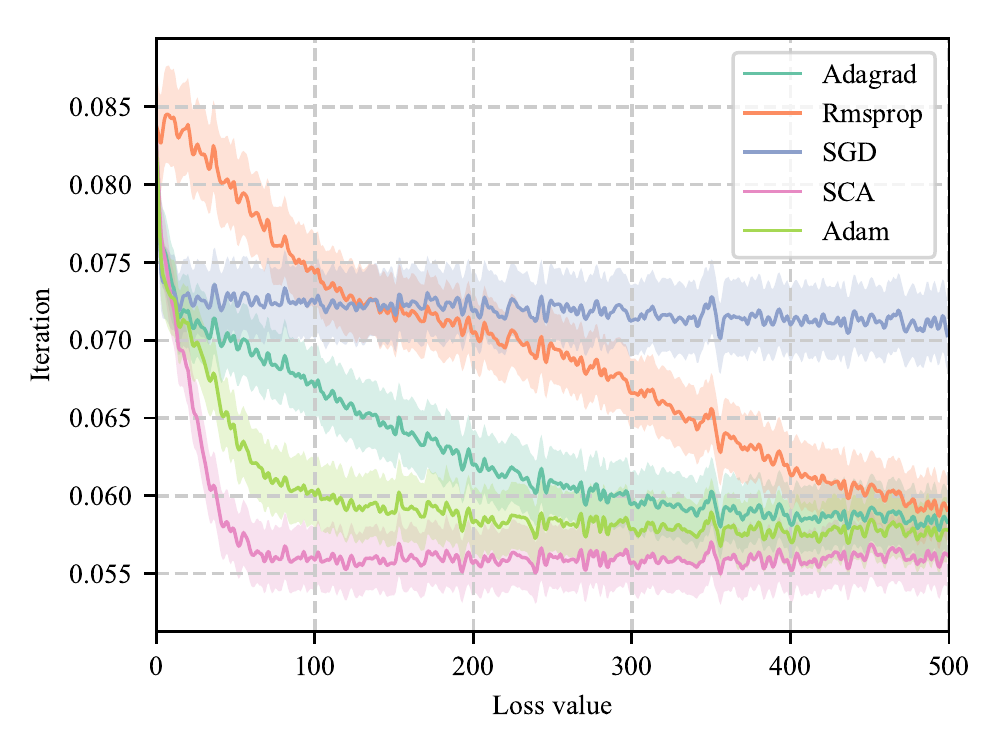}%		
	\label{fig:conv_wine}}
	\vfill
	\caption{Cost function value (per iteration) on the four datasets. The solid lines are the mean across runs, the shaded regions represent $\pm$ one standard deviation.}
	\label{fig:conv}
\end{figure*}

Results in terms of overall loss value per iteration are shown in Fig. \ref{fig:conv}, where the proposed algorithm is shown with a purple line, and the shaded areas represent the standard deviation around the mean. It can be seen that the overall results are generally consistent across the four datasets. Specifically, SGD is a relatively poor choice, getting stuck in a close minimum two out of four times (CASP and Wine datasets), and converging slowly in the other two cases (Parkinson and Skills). Among the state-of-the-art algorithms, Adam is always the best one (which is in accordance with its popularity), closely followed by Adagrad, with RMSProp scoring in the middle between these two and SGD. However, in all situations the proposed SCA technique is able to significantly outperform the other algorithms, being consistently faster in two out of four cases (Skills and Wine), and converging to a better solution in all settings.

\begin{table*}
\caption{Results (in terms of mean-squared error on the test set) of different algorithms. Best results for both groups are highlighted in bold.}
\setlength{\tabcolsep}{6pt}
\renewcommand{\arraystretch}{1.3}
\centering
\begin{small}
\begin{tabular}{lccccc}   %@{}p{0.3\columnwidth}p{0.2\columnwidth}p{0.2\columnwidth}@{}
\toprule
\textbf{Dataset} & SGD & AdaGrad & RMSProp & Adam & \textbf{Proposed} \\
\midrule
CASP & $0.2713 \pm 0.0053$ & $0.2348 \pm 0.0074$ & $0.2182 \pm 0.0114$ & $0.2107 \pm 0.0103$ & $\vect{0.2017 \pm 0.0045}$ \\
Parkinsons & $0.1550 \pm 0.0062$ & $0.1389 \pm 0.0042$ & $0.1417 \pm 0.0075$ & $0.1403 \pm 0.0058$ & $\vect{0.1374 \pm 0.0045}$ \\
SkillCraft1 & $0.0708 \pm 0.0040$ & $0.0688 \pm 0.0029$ & $0.0738 \pm 0.0083$ & $0.0707 \pm 0.0061$ & $\vect{0.0675 \pm 0.0034}$ \\
Wine & $0.0692 \pm 0.0029$ & $0.0551 \pm 0.0024$ & $0.0562 \pm 0.0025$ & $0.0543 \pm 0.0034$ & $\vect{0.0528 \pm 0.0023}$ \\
\bottomrule
\end{tabular}
\end{small}
\label{tab:results_regression_problems}
\end{table*}

The better results in term of convergence are equivalently found when considering the MSE on the independent test set, as shown in Table \ref{tab:results_regression_problems}. It can be seen that the final MSE for the SCA training algorithm always outperforms the MSE for the architectures optimized by competing algorithms. Considering the training time, we note that for these algorithms all algorithms required (approximately) the same training time per iteration, which is not shown here for reasons of space.

An interesting question is motivating theoretically the improvement in convergence time provided by the SCA methodology. To this end, one can observe that the matrix \eqref{eq:Ai} is in fact an approximation to the true Hessian matrix of the squared cost, which is obtained by assuming that the error is uncorrelated with the second derivative of the squared loss function. This is known as an outer-product approximation, or Levenberg-Marquardt approximation (see \cite[Section 5.4.2]{bishop2006pattern} for a discussion). Thus, it is reasonable to assume that the overall algorithm is able to maintain some information on the curvature of the cost function, even if higher-order derivatives with respect to the gradient are never explicitly computed. A similar argument was made in \cite{scardapane2017framework}. This is an interesting line of reasoning which could eventually lead to improved approximations for the cost function. 
\begin{figure}[t]
\centering
\includegraphics[width=0.45\columnwidth]{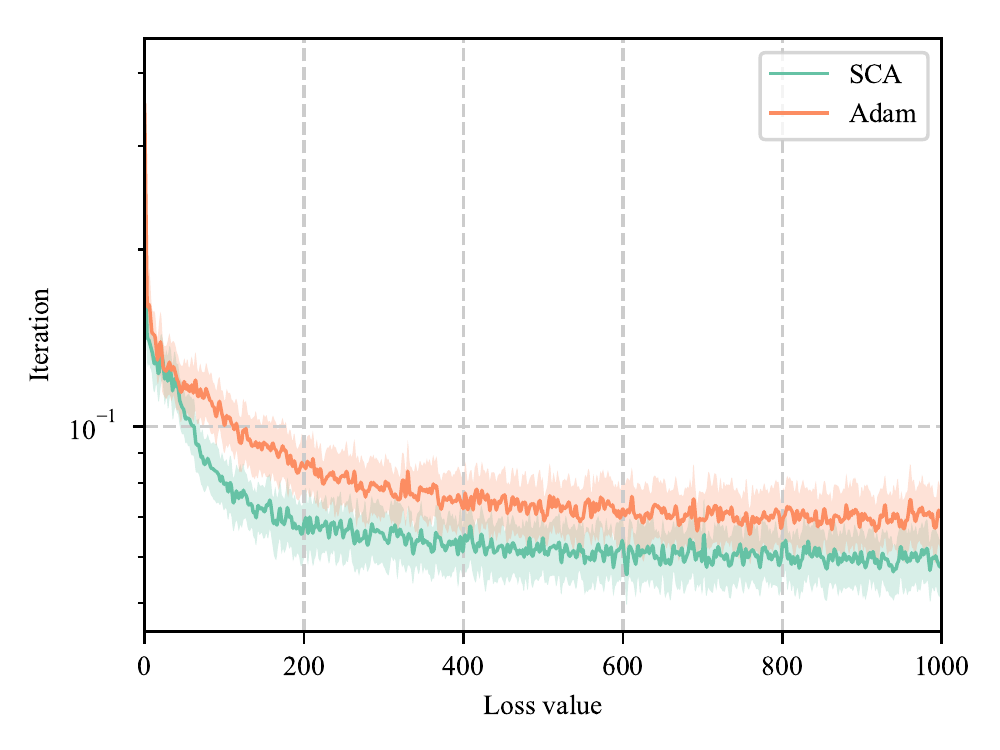}%
\caption{Cost function value (per iteration) on the Susy datasets. The solid lines are the mean across runs, the shaded regions represent $\pm$ one standard deviation.}
\label{fig:conv_susy}
\end{figure}
\subsection{Experiment on a large-scale dataset}
\label{sec:results_large_scale_dataset}
\begin{figure}[t]
\centering
\includegraphics[width=0.45\columnwidth]{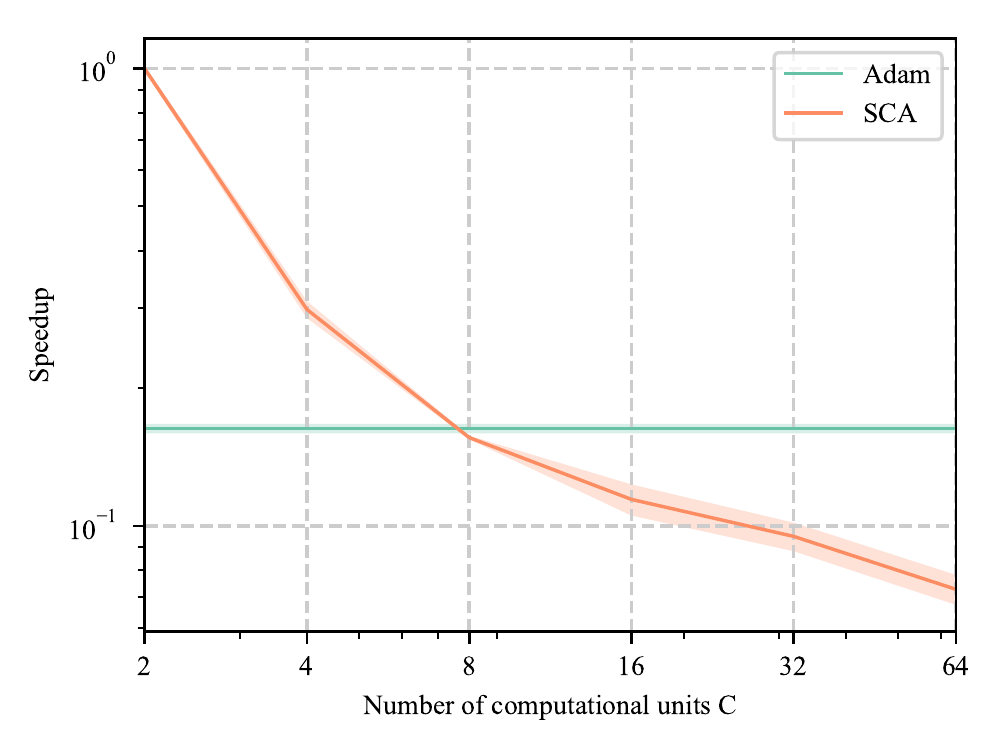}%
\caption{Relative speedup of the SCA procedure on the Susy dataset, when increasing $C$ from $2$ to $64$. The average training time for Adam is shown for a comparison. The solid lines are the mean across runs, the shaded regions represent $\pm$ one standard deviation.}
\label{fig:speedup_susy}
\end{figure}
Next, we consider a large-scale dataset to evaluate the performance of the algorithm on high-dimensional NNs. Due to the size of the parameter vector, this also allows us to test the parallel version of the algorithm discussed in Section \ref{sec:parallelizing_surrogate_optimization}. To this end, we consider the SUSY benchmark originally introduced in \cite{baldi2014searching}.\footnote{\url{https://archive.ics.uci.edu/ml/datasets/SUSY}} The dataset is composed by $5$ million simulations of particle collisions at high energy, simulating the environment found in currently used particle accelerators. The task is to distinguish between processes where supersymmetric particles are created (denoted as $\chi^{\pm}$ and $\chi^0$), and background processes. The challenge is that supersymmetric particles are not observed, and in both cases the observed particles are identical (leptons). Each example is described by an $18$-dimensional input vector, where the first $8$ features are low-level features describing the measurements, while the remaining $10$ features are high-level features constructed from the low-level ones. For this experiment, we use larger mini-batches of $L=50$ elements. The network has two hidden layers with $100$ neurons each, and we use a slightly larger regularization factor $\lambda=0.01$.

In Fig. \ref{fig:conv_susy} we show the convergence per iteration of Adam as compared to the proposed approach. The initial learning rates for the two algorithms were fine-tuned to obtain the fastest convergence behavior. Also, we use a parallel version with $C=4$, whose training time is roughly twice with respect to Adam, which is acceptable. It can be seen that, as for the previous section, the SCA technique significantly outperforms the Adam approach. The average test accuracy is similarly larger: on an independently kept $25\%$ of the original dataset, the network trained via the SCA algorithm has an average area under the curve (AUC) of $0.85$, which is eight points higher than the one obtained by Adam ($0.77$).

\begin{figure}[t]
\centering
\includegraphics[width=0.45\columnwidth]{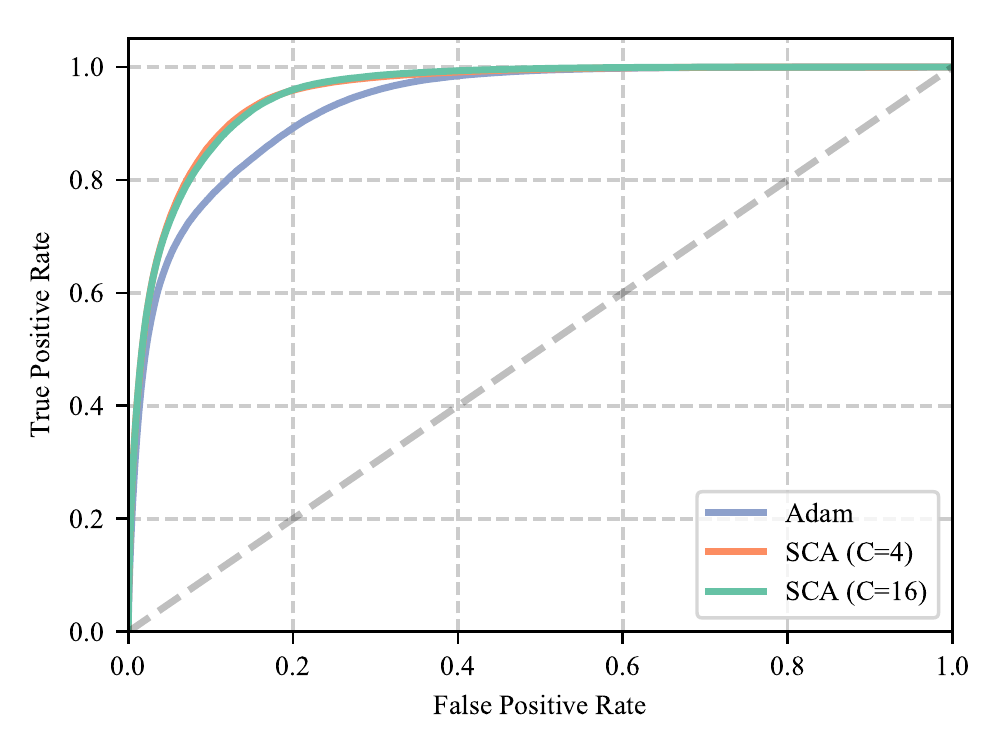}%
\caption{ROC curves shown for Adam and the proposed approach, with two different settings for $C$. The black line corresponds to random guessing.}
\label{fig:roc_curve_susy}
\end{figure}

To visualize the speedup obtained by the parallelization procedure, in Fig. \ref{fig:speedup_susy} we show the relative speedup with respect to $C=1$ obtained when varying $C$ in $2, 4, 8, \ldots, 64$. As can be shown, the training time for $C=2$ is generally higher than Adam, but the gap closes substantially for $C=4$. The two algorithms have almost the same training time for $C=8$, while SCA requires less than half the training time of Adam when having available $C=64$ computational units. This speedup is obtained without sacrificing accuracy. In Fig. \ref{fig:roc_curve_susy}, we show the receiver operating characteristic (ROC) curves obtained by Adam, SCA with $C=4$, and SCA with $C=16$. We can see that SCA outperforms the Adam algorithm (in line with our previous discussion). Additionally, the same ROC curve is obtained for both configurations of the method.

\section{Conclusions}
\label{sec:conclusions}

In this paper, we proposed a novel family of algorithms designed for optimizing neural networks. At each iteration of the algorithm, a random mini-batch of data is extracted from the training set, and a strongly convex approximation of the original training algorithm is solved. The algorithm only requires first-order information on the network. Our experimental results show that it performs favorably with respect to state-of-the-art approaches, being in general faster to converge to a better minimum of the optimization problem. For large-scale problems, the algorithm can be easily parallelized across multiple computational units. Further research will investigate the possibility of designing better approximation functions, and the customization of the framework to different families of NNs, including convolutional and recurrent networks. Additionally, we plan to test the algorithm on a large-scale cluster environment with asynchronous updates.

\bibliographystyle{IEEEtran}
\bibliography{biblio}

\vfill
\end{document}